\documentclass[]{iscram}

\usepackage[utf8]{inputenc}

\usepackage{lipsum}
\usepackage{tikz}
\usepackage{fancyvrb}
\usepackage{listings}
\usepackage{enumitem}
\usepackage{tcolorbox}
\usepackage{multicol}
\usepackage{siunitx}
\usepackage{xpatch}
\usepackage{svg}
\usepackage{multirow}
\usepackage{CJKutf8}

\lstdefinestyle{common}{
  xleftmargin=.5em,
  xrightmargin=.5em,
  frame=single,framesep=.5em,framerule=0pt,
  fancyvrb=true,
  basicstyle=\ttfamily,
  keywordstyle=\color{cyan!50!blue!75!black}\bfseries,
  commentstyle=\color{red!50!black}\itshape,
  stringstyle=\ttfamily\color{green!50!black},
  numbers=none,
  showspaces=false,
  showstringspaces=false,
  fontadjust=true,
  keepspaces=true,
  flexiblecolumns=true,
  emphstyle=\color{red},
}
\lstdefinestyle{TeX}{
  style=common,
  backgroundcolor=\color{blue!5},
  aboveskip=5pt,
  belowskip=5pt,
  language=[LaTeX]TeX,
  moretexcs={
    abstract, addbibresource, iscramset, keywords, mainmatter,
    maketitle, printbibliography, subsection, subsubsection, url,
    urldef, href, includegraphics, ldots, parencite, citeauthor,
    citeyear, citetitle, midrule, toprule, bottomrule
  },
  fancyvrb=true,
}
\lstdefinestyle{console}{
  style=common,
  backgroundcolor=\color{gray!10},
  aboveskip=5pt,
  belowskip=5pt,
}

\newlist{options}{description}{1}
\setlist[options]{%
  beginpenalty=10000,%
  itemsep=.5\parskip plus .3\parskip minus .2\parskip,
  parsep=.5\parskip plus .3\parskip minus .2\parskip,
  topsep=.5\parskip plus .3\parskip minus .2\parskip,
  partopsep=.5\parskip plus .3\parskip minus .2\parskip,
  style=nextline,labelindent=1em,%
  font=\normalfont\ttfamily}

\colorlet{macro color}{cyan!50!blue!75!black}
\colorlet{option color}{red!50!black}
\colorlet{generic color}{green!40!black}

\newtcolorbox{pseudoTeX}{colback=blue!5,colframe=blue!5,before=\nobreak}
\let\LaTeXorig\LaTeX
\renewcommand\LaTeX{\bgroup\fontfamily{lmr}\selectfont\upshape\LaTeXorig\egroup}

\urldef{\sitewebgind}\url{http://gind.mines-albi.fr}
\urldef{\sitewebminesalbi}\url{http://www.mines-albi.fr}
\urldef{\gmailpaulgaborit}\url{email adress}
\urldef{\jdmail}\url{...@example.com}
\urldef{\sitex}\url{www.example.com}

\iscramset{
  CoRe Paper 2024={Social Media for Crisis Management},
  title={Semantically Enriched Cross-Lingual Sentence Embeddings for Crisis-related Social Media Texts},
  short title={Cross-Lingual Sentence Embeddings for Crisis Texts},
  author={
    short name={Lamsal},
    full name=Rabindra Lamsal\thanks{corresponding author},
    affiliation={
      The University of Melbourne\\
       \href{mailto:rlamsal@student.unimelb.edu.au}{rlamsal@student.unimelb.edu.au}
    },
  },
  author={
    full name= Maria Rodriguez Read,
    affiliation={
      The University of Melbourne\\
       \href{mailto:maria.read@unimelb.edu.au}{maria.read@unimelb.edu.au}
    },
  },
  author={
    full name= Shanika Karunasekera,
    affiliation={
      The University of Melbourne\\
       \href{mailto:karus@unimelb.edu.au}{karus@unimelb.edu.au}
    },
  },
}

\usepackage{tikzpagenodes}
\usetikzlibrary{fit}

\begin{document}

\maketitle


\abstract{
Tasks such as semantic search and clustering on crisis-related social media texts enhance our comprehension of crisis discourse, aiding decision-making and targeted interventions. Pre-trained language models have advanced performance in crisis informatics, but their contextual embeddings lack semantic meaningfulness. Although the CrisisTransformers family includes a sentence encoder to address the semanticity issue, it remains monolingual, processing only English texts. Furthermore, employing separate models for different languages leads to embeddings in distinct vector spaces, introducing challenges when comparing semantic similarities between multi-lingual texts. Therefore, we propose multi-lingual sentence encoders (\texttt{CT-XLMR-SE} and \texttt{CT-mBERT-SE}) that embed crisis-related social media texts for over 50 languages, such that texts with similar meanings are in close proximity within the same vector space, irrespective of language diversity. Results in sentence encoding and sentence matching tasks are promising, suggesting these models could serve as robust baselines when embedding multi-lingual crisis-related social media texts. The models are publicly available at: \texttt{\url{https://huggingface.co/crisistransformers}}.
}


\section{Introduction}
In times of crisis, social media platforms such as Facebook and Twitter are critical channels for information sharing and communication (\cite{imran2015processing,simon2015socializing,lamsal2022socially}). These platforms help promptly disseminate essential information, whether it is related to wildfires, earthquakes, hurricanes, floods, epidemics, etc (\cite{starbird2010pass,thomson2012trusting,alam2018twitter,pourebrahim2019understanding,lamsal2023twitter}). They serve as central information hubs for both the general population and emergency responders, providing updates on unfolding situations (\cite{sarcevic2012beacons}). Social media also assists in allowing individuals to seek and offer assistance while coordinating relief efforts (\cite{purohit2014emergency}). The extensive amount of user-generated content on these platforms is a valuable historical and real-time data source. However, the challenges of analyzing and understanding crisis-related social media texts arise from their sheer volume (\cite{stieglitz2018social}) and linguistic complexities. As the number of conversations exponentially increases during a crisis, automated analysis becomes necessary to comprehend the situation at the ground level. Such conversations contain situational information (\cite{hughes2009twitter,vieweg2010microblogging,vieweg2012situational}) about the affected population, damages, injuries, casualties, rescue and volunteering efforts, etc. (\cite{imran2016twitter}), shared by populations, whether directly or indirectly impacted (\cite{imran2015processing,lamsal2022socially}).

The majority of analyses in processing crisis-related social media texts involve tasks like text classification (\cite{ashktorab2014tweedr,caragea2011classifying,imran2013extracting,li2018comparison}), semantic search (\cite{dutt2019utilizing}), and clustering (\cite{ashktorab2014tweedr,curiskis2020evaluation,lamsal2022socially}). The advancement in these areas plays a key role in enhancing our comprehension of crisis discourse to facilitate informed decision-making processes and assist in the development of targeted interventions and communication strategies. This holds true not only during the course of an unfolding crisis but also retrospectively, drawing insights from historical events. The knowledge extracted from previous crises significantly contributes to the formulation of effective and efficient response strategies for similar future crisis scenarios, thus strengthen preparedness and response capabilities for future challenges.

Transformer-based (\cite{vaswani2017attention}) pre-trained language models like BERT (\cite{devlin2018bert}) and RoBERTa (\cite{liu2019roberta}) have significantly advanced performance in numerous NLP tasks across domains including crisis informatics. Moreover, the recently introduced CrisisTransformers (\cite{lamsal2023crisistransformers}), an ensemble of pre-trained models trained on a corpus of over 15 billion word tokens from more than 30 crisis events, has further improved the state-of-the-art. However, the contextual embeddings provided by these pre-trained models lack semantic meaningfulness\footnote{Semantically similar sentences are embedded together in a vector space, and those embeddings can be compared using similarity measures such as cosine similarity.} and perform worse than averaging GloVe embeddings (\cite{reimers2019sentence}). For semantic search and clustering tasks, it is crucial to have sentence embeddings with semantic richness. While the CrisisTransformers family does offer a sentence encoder for the task, improving upon the Sentence Transformers (\cite{reimers2019sentence}), it remains monolingual, processing only English-language social media texts. Any geographical region can have a diverse linguistic population, whether it's a county, city, state, or country. Social media platforms, therefore, can be flooded with posts in various languages during the same crisis event. Analyzing solely the texts in a specific language increases the risk of overlooking essential information available in texts shared in other languages. Also, employing separate models for different languages leads to embeddings in distinct vector spaces. This discrepancy poses a challenge when comparing the semantic similarities between sentences in various languages. 

\begin{figure}
    \centering
    \includegraphics[width=0.6\textwidth]{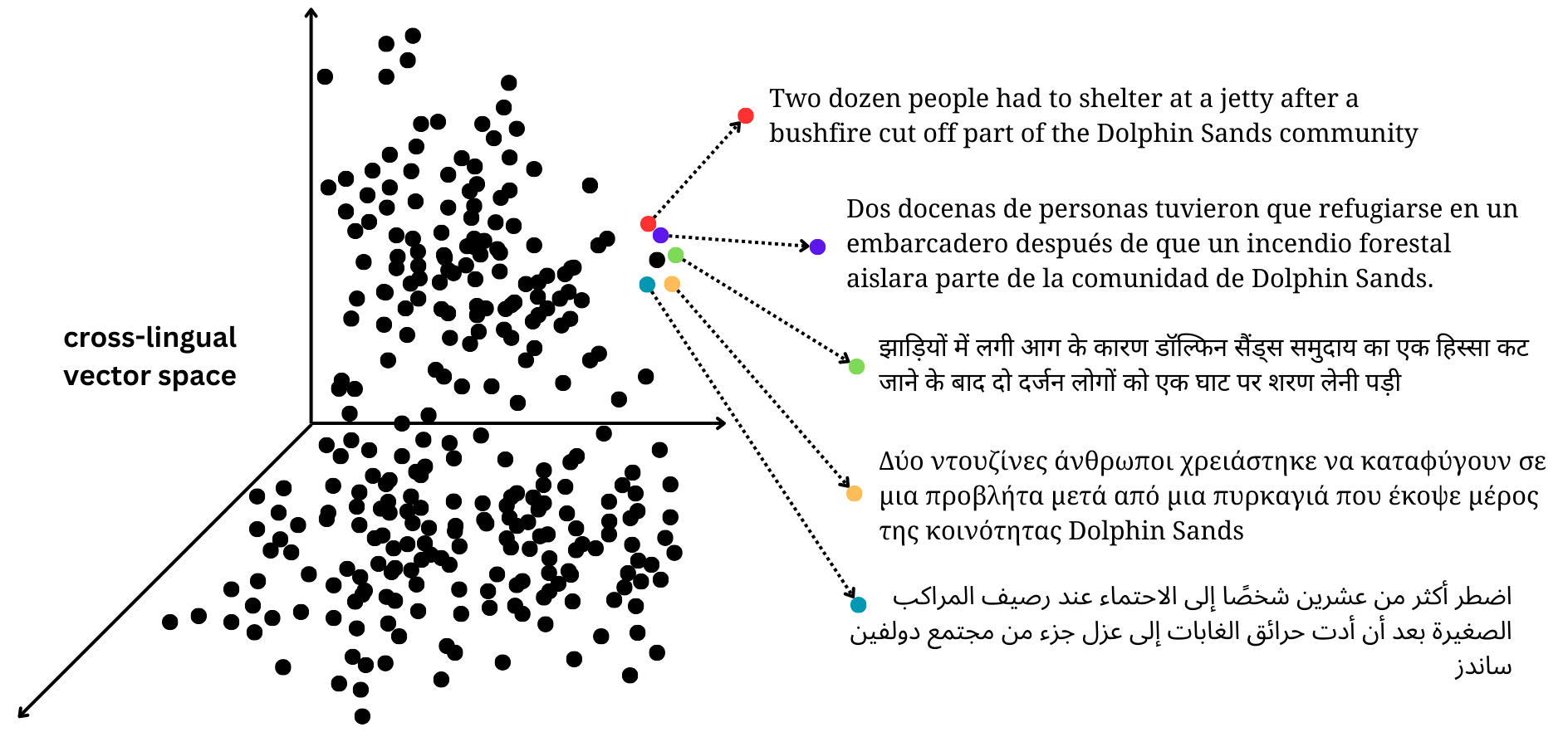}
    \caption{An illustration of a cross-lingual vector space for crisis-related social media texts.}
    \label{vector-space}
\end{figure}

Adopting a single embedding space is critical to addressing these challenges and enhancing the effectiveness of crisis informatics. Such an encoder would facilitate mapping texts from different languages with similar meanings to close proximity within the same vector space, as illustrated in Figure \ref{vector-space}. This approach ensures that semantically related content in diverse languages can be effectively processed, enabling a more comprehensive and nuanced understanding of crisis-related information irrespective of linguistic diversity. Therefore, we address this need by introducing the first-ever multi-lingual sentence encoders for embedding crisis-related social media texts. In general, this study contributes the following to the existing crisis informatics literature:

\begin{itemize}
    \item We introduce two multi-lingual sentence encoders (\texttt{CT-XLMR-SE\footnote{\texttt{\url{https://huggingface.co/crisistransformers/CT-XLMR-SE}}}} and \texttt{CT-mBERT-SE\footnote{\texttt{\url{https://huggingface.co/crisistransformers/CT-mBERT-SE}}}}) that embed crisis-related social media texts with semantic richness for 52 languages: Albanian, Arabic, Armenian, Bulgarian, Catalan, Chinese (Simplified), Chinese (Traditional), Croatian, Czech, Danish, Dutch, Estonian, Finnish, French, French (Canada), Galician, Georgian, German, Greek, Gujarati, Hebrew, Hindi, Hungarian, Indonesian, Italian, Japanese, Korean, Kurdish (Sorani), Latvian, Lithuanian, Macedonian, Malay, Marathi, Mongolian, Myanmar (Burmese), Norwegian, Persian, Polish, Portuguese, Portuguese (Brazil), Romanian, Russian, Serbian, Slovak, Slovenian, Spanish, Swedish, Thai, Turkish, Ukrainian, Urdu, and Vietnamese.
    \item We publicly release the sentence encoders, making them easily accessible for integration with the Transformers library (\cite{wolf2020transformers}). We anticipate that these sentence encoders will serve as robust baselines for tasks that involve embedding multi-lingual crisis-related social media texts.
\end{itemize}

\section{Related Work}
Semantically rich embeddings position similar sentences closely together in a vector space. Learning such embeddings is an extensively explored area, briefly discussed in this section. (\cite{kiros2015skip}) trained an encoder-decoder model to reconstruct the neighbouring sentences of an encoded sequence. This training aimed to map sentences with similar semantic properties to comparable vector representations. (\cite{conneau2017supervised}) introduced a siamese Bidirectional Long Short-Term Memory (BiLSTM) network, incorporating max-pooling on the Stanford Natural Language Inference (SNLI) dataset. This approach surpassed previous unsupervised methods (\cite{kiros2015skip,hill2016learning}). (\cite{cer2018universal}) extended unsupervised learning by training a transformer network on the SNLI dataset. Furthermore, (\cite{yang2018learning}) proposed an unsupervised learning method for sentence-level semantic similarity based on conversational data. 

With the introduction of BERT in 2018, the unsupervised training component was replaced with pre-trained models in designing sentence encoders. (\cite{reimers2019sentence}) fine-tuned BERT using siamese and triplet networks on the SNLI and Multi-Genre Natural Language Inference (MultiNLI) datasets. The fine-tuning involved a softmax classifier with ``contradiction", ``entailment", and ``neutral" labels. Similarly, (\cite{gao2021simcse}) proposed SimCSE, a contrastive approach to fine-tune pre-trained models using natural language inference datasets, utilizing ``contradiction" pairs as hard negatives. Building on this, (\cite{reimers2019sentence}) fine-tuned multiple pre-trained models using over 1 billion sentence pairs and released the second version of their models. Following the approach of utilizing ``contradiction" pairs as hard negatives, as proposed in (\cite{gao2021simcse}), (\cite{lamsal2023crisistransformers}) replaced pre-trained models such as BERT and RoBERTa, which were trained on texts from broad and general domains with crisis-specific pre-trained models. Their domain-specific models, publicly released as CrisisTransformers, were trained on an extensive corpus containing over 15 billion word tokens from 30+ crisis events, including disease outbreaks, natural disasters, protests and activism, conflicts, civil war, etc. Performance improvement of $>$17\% was reported compared to Sentence Transformers (\cite{reimers2019sentence}) in sentence encoding tasks for crisis-related social media texts. However, their sentence encoder is monolingual, processing only English language texts.

Multiple approaches have been proposed in the literature to train multi-lingual sentence embeddings. (\cite{artetxe2019massively}) trained an encoder-decoder network on parallel corpora with a translation task and used output from the encoder as the sentence embedding. While this approach is effective in identifying exact translations across various languages, its performance diminishes when evaluating the similarity of sentences that are not exact translations. (\cite{yang2019multilingual}) learnt multi-lingual sentence embeddings using a multi-task setup on the SNLI dataset and millions of question-answer pairs. They used a translation ranking task to align cross-lingual vector spaces. However, this approach is susceptible to catastrophic interference and has significant computational overhead. In addressing these issues, (\cite{reimers2020making}) proposed a training approach (Sentence Transformers) to map a translated sentence to the same location in the vector space as the original sentence. In this way, a multi-lingual model (student model) can be trained on translation pairs to mimic a mono-lingual model (teacher model) and learn a common vector space for multiple languages. In this study, we consider Sentence Transformer’s top-performing model \textit{all-mpnet-base-v2} as a baseline.

\section{Materials and methods}

\subsection{Training architecture}

\begin{figure}
    \centering
    \includegraphics[width=\textwidth]{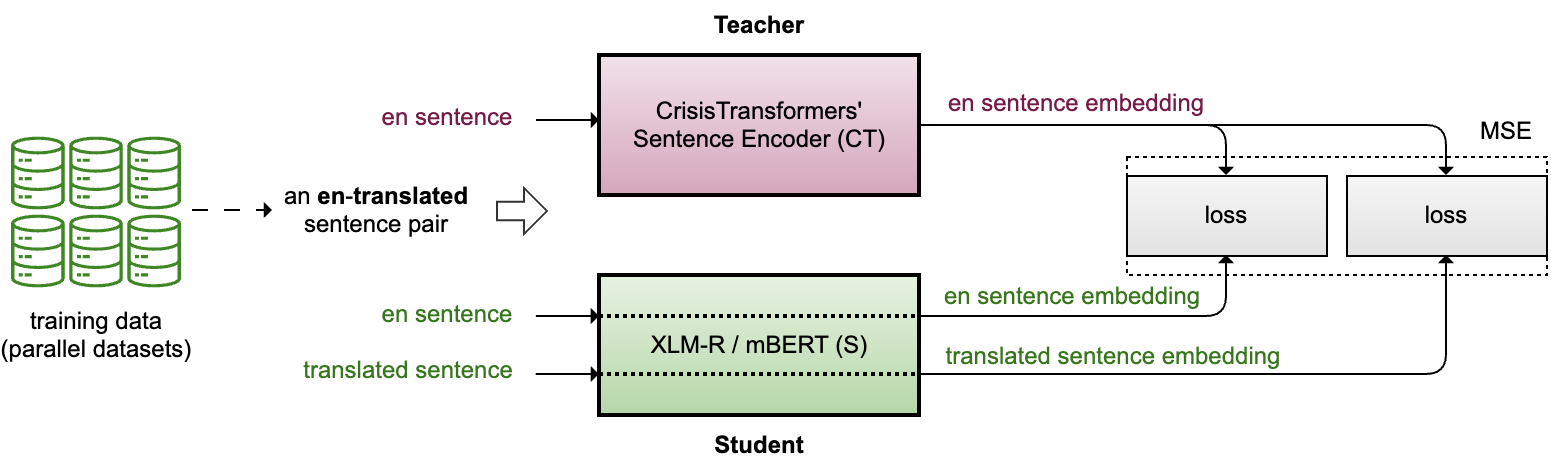}
    \caption{The student-teacher training architecture.}
    \label{multilingual-training}
\end{figure}

We utilized CrisisTransformers' sentence encoder (\cite{lamsal2023crisistransformers}) as a teacher model and extended it to develop multi-lingual models with a student-teacher training architecture (\cite{reimers2020making}). The training network is illustrated in Figure \ref{multilingual-training}. We considered XLM-R (\cite{conneau2019unsupervised}) and mBERT (\cite{devlin2018bert}) as student models. XLM-R is the multi-lingual version of RoBERTa, trained on 2.5TB of CommonCrawl data containing 100 languages. Similarly, mBERT is the multi-lingual version of BERT, trained on Wikipedia data for 104 languages.

\textbf{Training}: For a dataset $D$ = (($en_1$,$t_1$), ($en_2$,$t_2$), ..., ($en_n$,$t_n$)), where $en_i$ are English sentences and $t_i$ are respective translated sentences (can be any language), we trained a network with student model $\mathcal{S}$ and teacher model $\mathcal{CT}$ while minimizing the following training objective for a given mini-batch $\mathcal{B}$:

\begin{equation}
    \frac{1}{|\mathcal{B}|} \sum_{j \in \mathcal{B}} \left[ (\mathcal{CT}(en_j) - \mathcal{S}(en_j))^2 +  (\mathcal{CT}(en_j) - \mathcal{S}(t_j))^2 \right]
\end{equation}

With the above training objective, trained on dataset $D$ we aim to obtain $\mathcal{S}$ such that $\mathcal{S}(en_i)$ $\approx$ $\mathcal{CT}(en_i)$ and $\mathcal{S}(t_i)$ $\approx$ $\mathcal{CT}(en_i)$. $\mathcal{S}$ will be our multi-lingual model. We train XLM-R and mBERT in separate networks with common teacher $\mathcal{CT}$, and their final checkpoints are named \texttt{CT-XLMR-SE} and \texttt{CT-mBERT-SE} (where \textit{SE} stands for \textit{sentence encoder}). The training was done on an NVIDIA A100 GPU (80GB) for a maximum of 20 epochs, and 20k steps were used to warm up the learning rate with AdamW as an optimizer. Mixed precision training was done to improve the training time. We implemented the following training setting: maximum sequence length: 128, batch size: 64, maximum sentence per training file: 500k, and learning rate: 2e-5. The training of both networks in parallel finished in approximately 30 days.

\subsection{Training Data}

The training data $\mathcal{D}$ comprised multiple publicly available parallel datasets merged together. For each language, the maximum number of sentence pairs considered for training from any single dataset was 500k. Some samples in certain datasets had a single entry, either a sentence in English or any other language; in such cases, those samples were ignored. In total, the training data included over 128 million sentence pairs. The released versions of all the datasets already had the texts pre-processed. Some example pairs from the training data are provided in Table \ref{sample}. The following parallel datasets were considered in this study:

\begin{itemize}
    \item \textbf{Europarl} (\cite{koehn2005europarl}): Sentences extracted from the proceedings of the European Parliament.
    \item \textbf{GlobalVoices} (\cite{tiedemann2012parallel}): News stories collected from the website globalvoices.org.
    \item \textbf{JW300} (\cite{agic2019jw300}): Publications crawled from the website jw.org.
    \item \textbf{MUSE} (\cite{conneau2017word}): Ground-truth bilingual dictionaries.
    \item \textbf{News-Commentary} (\cite{barrault2019findings}): News commentaries collected from the website Project Syndicate, provided by WMT for shared tasks.
    \item \textbf{OpenSubtitles} (\cite{lison2016opensubtitles2016}): Large-scale collection of movie and TV subtitles.
    \item \textbf{Tatoeba}\footnote{https://tatoeba.org/}: Large-scale collection of sentences and their translations.
    \item \textbf{TED2020} (\cite{reimers2020making}): Translated subtitles of over 4,000 TED talks.
    \item \textbf{WikiMatrix} (\cite{schwenk2019wikimatrix}): Parallel sentences from Wikipedia's contents.
    \item \textbf{WikiTitles} (\cite{reimers2020making}): Cross-lingual Wikipedia titles extracted from Wikipedia dumps.
\end{itemize}

\begin{table}[]
    \centering
    \caption{Example sentence pairs.}
    \label{sample}

    \begin{tabular}{p{7.5cm}|p{7.5cm}}
        \toprule
      \textbf{english sentence (en)} & \textbf{translated sentence (t)}  \\
      \midrule
      \midrule
      1917 also saw the last issuance of 2 kroner coins. & [Portuguese] 1917 também viu a última emissão de moedas de 2 kroner. \\
      \hline
       Resumption of the session.  & [Spanish] Reanudación del período de sesiones. \\   
       \hline
       And then we can ask questions, real questions, questions like, what's the best life insurance policy to get? -- real questions that people have in their everyday lives. & [German] Und dann können wir Fragen stellen, echte Fragen, Fragen wie: Welche ist die beste Lebensversicherung? – echte Fragen, die die Leute im täglichen Leben haben. \\
       \hline
        Then they surrendered and so did the 4 other ships, one after another.	& [Dutch] Toen gaven ze zich over, evenals de vier andere schepen, de een na de ander.\\
        \hline
        Declaration of Fundamental Rights and Freedoms adopted by the Parliament.	& [French] Déclaration des droits et libertés fondamentales adopté par le Parlement.\\
        \hline 
        Most economists failed to foresee the economic dynamics that actually led to the crisis, because they failed to pay enough attention to the rapid increase in US total debt. & [Italian]	La maggior parte degli economisti non riuscì a prevedere le dinamiche economiche che condussero davvero alla crisi, perché non prestò sufficiente attenzione al rapido incremento del debito complessivo americano.\\
         \bottomrule
    \end{tabular}
\end{table}

\subsection{Evaluations}

\subsubsection{Sentence encoding tasks}
We implemented the sentence encoding task described in (\cite{lamsal2023crisistransformers}) to evaluate the embeddings generated by students compared to those by the teacher. This assessment measures how closely the students have replicated the teacher's vector space. For a crisis-related labelled dataset, we calculated the weighted average cosine similarity among tweets within individual classes as below:

\begin{equation}
    D_{\text{avg}} = \sum_{k=1}^{K} \hat{w}_k \cdot \frac{1}{|\{i : y_i = c_k\}|} \sum_{i : y_i = c_k} \text{similarity}(\mathbf{e}_i, \mathbf{e}_j)
\end{equation}

where, $K$ is the number of classes in a dataset, $\hat{w}_k$ is the normalized class weight, $\text{similarity}(\mathbf{e}_i, \mathbf{e}_j)$ computes the cosine similarity between embeddings $\mathbf{e}_i$ and $\mathbf{e}_j$, where $\mathbf{e}_j$ is a crisis-text from same class as $\mathbf{e}_i$.

\textbf{Datasets}: For this task, we utilized 18 publicly available human-labelled crisis-related texts datasets, all of which are in English. The following datasets were considered: CrisisMMD (\cite{alam2018crisismmd}), CrisisLex (\cite{olteanu2014crisislex}), AIDR (\cite{imran2014aidr}), ISCRAM2013 (\cite{imran2013extracting}), SWDM2013 (\cite{imran2013practical}), CrisisNLP (\cite{imran2016twitter}), COVID-19 stance (\cite{poddar2022winds}), SAD Stressor (\cite{mauriello2021sad}), SAD Stress (\cite{mauriello2021sad}), SAD COVID (\cite{mauriello2021sad}), LocBERT (\cite{lamsal2022did}), HMC (a) (\cite{biddle2020leveraging}), Vaccination opinions (\cite{cotfas2021longest}), HMC (b) (\cite{biddle2020leveraging}), PHM (\cite{karisani2018did}), COVID-19 patients and COVID-19 contacts (\cite{klein2021toward}), and ANTiVax (\cite{hayawi2022anti}).

\subsubsection{Sentence matching tasks}

Next, we assessed the multi-lingual aspect of the student models. We considered the test sets provided in TED2020 (\cite{reimers2020making}) to evaluate our student models. Each test set comprised 1,000 sentence pairs. For every test set and each student, we performed sentence matching tasks as follows:

Let $E$ be the set of English sentences and $T$ be the set of translated sentences. We have $n$ sentence pairs, where \(E = \{en_1, en_2, ..., en_n\}\) and \(T = \{t_1, t_2, ..., t_n\}\). For each $i$ (where \(1 \leq i \leq n\)), we compute the embeddings $\mathcal{S}(en_i)$ and $\mathcal{S}(t_i)$ for the English and translated sentences, respectively. 

\textbf{English to translated language evaluation}: For each $i$, assess if $t_i$ has the highest cosine similarity with $en_i$ compared to all other translated sentences in the set $T$:

\begin{equation}
    \text{en-t-match}_i = \begin{cases} 1 & \text{similarity}(\mathcal{S}(en_i), \mathcal{S}(t_i)) \geq \text{similarity}(\mathcal{S}(en_i), \mathcal{S}(t_j)) \text{ for all } t_j \neq t_i \\ 0 & \text{else} \end{cases}
    \label{en-t}
\end{equation}

\textbf{Translated language to English evaluation}: For each $i$, assess if $en_i$ has the highest cosine similarity with $t_i$ compared to all other English sentences in the set $E$:

\begin{equation}
    \text{t-en-match}_i = \begin{cases} 1 & \text{similarity}(\mathcal{S}(t_i), \mathcal{S}(en_i)) \geq \text{similarity}(\mathcal{S}(t_i), \mathcal{S}(en_j)) \text{ for all } en_j \neq en_i \\ 0 & \text{else} \end{cases}
    \label{t-en}
\end{equation}

In Equations \ref{en-t} and \ref{t-en}, ``1" denotes a correct match, and ``0" an incorrect match. The accuracy for each language is calculated by dividing the total number of correct matches by the number of sentence pairs in the test set.

\subsection{Sentence Embeddings}
For generating sentence embeddings, we performed mean pooling on the token embeddings of an input sentence while considering the attention mask. The following pre-processing steps were performed to each input sequence: (i) replacing URLs with the \texttt{HTTPURL} token, (ii) substituting mentions with the \texttt{@MENTION} token, (iii) decoding HTML entities, (iv) eliminating newline characters and unnecessary whitespaces, (v) fixing text encoding issues for consistency, and (vi) replacing emojis with their corresponding textual representations. A tweet was considered one input sequence with the maximum sequence length set to 128.

\section{Results and Discussions}

\begin{table}
\centering
\caption{Results (weighted average cosine similarity) from sentence encoding tasks.}
\label{eval-2}
\begin{tabular}{lc|ccc}
\toprule
\textbf{Dataset}                                         & \textbf{SBERT}  & \textbf{CT}     & \textbf{CT-XLMR-SE} & \textbf{CT-mBERT-SE} \\
\midrule
\midrule
CrisisMMD (\cite{alam2018crisismmd})            & 0.6103 & 0.7191 & 0.6772     & 0.6724      \\
CrisisLex (\cite{olteanu2014crisislex})         & 0.6809 & 0.8070  & 0.7705     & 0.7734      \\
AIDR (\cite{imran2014aidr})                     & 0.5407 & 0.6128 & 0.5586     & 0.5644      \\
ISCRAM2013 (\cite{imran2013extracting})         & 0.5632 & 0.6872 & 0.6545     & 0.6546      \\
SWDM2013 (\cite{imran2013practical})            & 0.4870  & 0.6744 & 0.6312     & 0.6314      \\
CrisisNLP (\cite{imran2016twitter})             & 0.5698 & 0.6960  & 0.6390      & 0.6446      \\
COVID-19 stance (\cite{poddar2022winds})        & 0.6019 & 0.8742 & 0.8444     & 0.8451      \\
SAD Stressor (\cite{mauriello2021sad})          & 0.7528 & 0.7027 & 0.6818     & 0.6864      \\
SAD Stress (\cite{mauriello2021sad})            & 0.8749 & 0.9021 & 0.8961     & 0.8974      \\
SAD COVID (\cite{mauriello2021sad})             & 0.6674 & 0.7167 & 0.7134     & 0.7147      \\
LocBERT (\cite{lamsal2022did})                  & 0.7297 & 0.8982 & 0.8750      & 0.8834      \\
HMC (a) (\cite{biddle2020leveraging})           & 0.8211 & 0.8342 & 0.8101     & 0.8138      \\
Vaccination opinions (\cite{cotfas2021longest}) & 0.4939 & 0.7874 & 0.7543     & 0.7633      \\
HMC (b) (\cite{biddle2020leveraging})           & 0.7259 & 0.7762 & 0.7336     & 0.7379      \\
PHM (\cite{karisani2018did})                    & 0.7579 & 0.7708 & 0.7478     & 0.7537      \\
COVID-19 patients (\cite{klein2021toward})      & 0.5320  & 0.6549 & 0.7054     & 0.7082      \\
COVID-19 contacts (\cite{klein2021toward})      & 0.5192 & 0.5874 & 0.6034     & 0.5984      \\
ANTiVax (\cite{hayawi2022anti})                   & 0.5450  & 0.7722 & 0.7631     & 0.7614   \\
\midrule
Average & 0.6374 & 0.7485 & 0.7255 & 0.7280 \\
\bottomrule
\end{tabular}
\end{table}

\begin{table}
    \centering
    \caption{Performance of vanilla XLM-R and mBERT on sentence encoding tasks.}
    \label{vanilla}
    \begin{tabular}{c|c|c}
    \toprule
         & \textbf{XLM-R} & \textbf{mBERT} \\
         \midrule
         \midrule
       Average  & 0.0032 & 0.3581\\
       \bottomrule
    \end{tabular}
\end{table}

\begin{table}
\centering
\caption{Results (accuracy) from sentence matching tasks. For each language, the best average score is highlighted and equal scores across models are underlined. \texttt{en-t}: English to translated language and \texttt{t-en} is translated language to English.}
\label{eval-1}

\begin{tabular}{p{3.8cm}|lll|lll}
\toprule
& \multicolumn{3}{c}{\textbf{CT-XLMR-SE}}   & \multicolumn{3}{|c}{\textbf{CT-mBERT-SE}} \\
\midrule
\textbf{Language}     & \textbf{en-t acc.} & \textbf{t-en acc.} & \textbf{avg.}   & \textbf{en-t acc.} & \textbf{t-en acc.} & \textbf{avg.}   \\
\midrule
\midrule
Albanian              & 0.9420        & 0.9480        & 0.9450          & 0.9510        & 0.9500        & \textbf{0.9505} \\
Arabic                & 0.9430        & 0.9440        & 0.9435          & 0.9420        & 0.9380        & \textbf{0.9400} \\
Armenian              & 0.9200        & 0.9100        & \textbf{0.9150} & 0.9170        & 0.9050        & 0.9110          \\
Bulgarian             & 0.9670        & 0.9660        & \textbf{0.9665} & 0.9650        & 0.9640        & 0.9645          \\
Catalan               & 0.9600        & 0.9610        & \textbf{0.9605} & 0.9520        & 0.9530        & 0.9525          \\
Chinese (Simplified)  & 0.9420        & 0.9430        & \textbf{0.9425} & 0.9410        & 0.9280        & 0.9345          \\
Chinese (Traditional) & 0.9420        & 0.9280        & 0.9350          & 0.9460        & 0.9260        & \textbf{0.9360} \\
Croatian              & 0.9710        & 0.9710        & 0.9710          & 0.9750        & 0.9700        & \textbf{0.9725} \\
Czech                 & 0.9610        & 0.9590        & 0.9600          & 0.9630        & 0.9630        & \textbf{0.9630} \\
Danish                & 0.9640        & 0.9640        & \textbf{0.9640} & 0.9610        & 0.9650        & 0.9630          \\
Dutch                 & 0.9760        & 0.9720        & \textbf{0.9740} & 0.9760        & 0.9700        & 0.9730          \\
Estonian              & 0.9340        & 0.9340        & \textbf{0.9340} & 0.9270        & 0.9220        & 0.9245          \\
Finnish               & 0.9390        & 0.9390        & 0.9390          & 0.9430        & 0.9370        & \textbf{0.9400} \\
French                & 0.9710        & 0.9680        & \textbf{0.9695} & 0.9730        & 0.9650        & 0.9690          \\
French (Canada)       & 0.9530        & 0.9490        & 0.9510          & 0.9570        & 0.9460        & \textbf{0.9515} \\
Galician              & 0.9630        & 0.9640        & \underline{0.9635}    & 0.9650        & 0.9620        & \underline{0.9635}    \\
Georgian              & 0.8890        & 0.9000        & 0.8945          & 0.9130        & 0.9010        & \textbf{0.9070} \\
German                & 0.9760        & 0.9710        & \textbf{0.9735} & 0.9740        & 0.9700        & 0.9720          \\
Greek                 & 0.9700        & 0.9700        & \textbf{0.9700} & 0.9660        & 0.9600        & 0.9630          \\
Gujarati              & 0.8600        & 0.8650        & 0.8625          & 0.8990        & 0.8860        & \textbf{0.8925} \\
Hebrew                & 0.9590        & 0.9610        & \textbf{0.9600} & 0.9580        & 0.9610        & 0.9595          \\
Hindi                 & 0.9060        & 0.9080        & \textbf{0.9070} & 0.8990        & 0.9010        & 0.9000          \\
Hungarian             & 0.9620        & 0.9580        & \textbf{0.9600} & 0.9600        & 0.9550        & 0.9575          \\
Indonesian            & 0.9740        & 0.9620        & \textbf{0.9680} & 0.9730        & 0.9610        & 0.9670          \\
Italian               & 0.9640        & 0.9600        & 0.9620          & 0.9650        & 0.9610        & \textbf{0.9630} \\
Japanese              & 0.9240        & 0.9150        & \textbf{0.9195} & 0.9240        & 0.9140        & 0.9190          \\
Korean                & 0.9450        & 0.9420        & \textbf{0.9435} & 0.9350        & 0.9310        & 0.9330          \\
Kurdish (Sorani)      & 0.8940        & 0.8840        & \textbf{0.8890} & 0.3460        & 0.2650        & 0.3055          \\
Latvian               & 0.9560        & 0.9550        & \textbf{0.9555} & 0.9550        & 0.9520        & 0.9535          \\
Lithuanian            & 0.9420        & 0.9310        & \textbf{0.9365} & 0.9310        & 0.9330        & 0.9320          \\
Macedonian            & 0.9790        & 0.9820        & \textbf{0.9805} & 0.9790        & 0.9810        & 0.9800          \\
Malay                 & 0.9190        & 0.9190        & \textbf{0.9190} & 0.9120        & 0.9100        & 0.9110          \\
Marathi               & 0.8300        & 0.8340        & 0.8320          & 0.8680        & 0.8580        & \textbf{0.8630} \\
Mongolian             & 0.8430        & 0.8300        & 0.8365          & 0.8730        & 0.8670        & \textbf{0.8700} \\
Myanmar (Burmese)     & 0.9200        & 0.9070        & \textbf{0.9135} & 0.9110        & 0.9010        & 0.9060          \\
Norwegian             & 0.9470        & 0.9460        & 0.9465          & 0.9520        & 0.9470        & \textbf{0.9495} \\
Persian               & 0.9290        & 0.9280        & \textbf{0.9285} & 0.9230        & 0.9250        & 0.9240          \\
Polish                & 0.9560        & 0.9460        & 0.9510          & 0.9560        & 0.9490        & \textbf{0.9525} \\
Portuguese            & 0.9700        & 0.9780        & \textbf{0.9740} & 0.9700        & 0.9770        & 0.9735          \\
Portuguese (Brazil)   & 0.9780        & 0.9850        & \textbf{0.9815} & 0.9770        & 0.9790        & 0.9780          \\
Romanian              & 0.9680        & 0.9700        & 0.9690          & 0.9750        & 0.9710        & \textbf{0.9730} \\
Russian               & 0.9610        & 0.9620        & \textbf{0.9615} & 0.9640        & 0.9570        & 0.9605          \\
Serbian               & 0.9750        & 0.9670        & \textbf{0.9710} & 0.9710        & 0.9690        & 0.9700          \\
Slovak                & 0.9730        & 0.9690        & \textbf{0.9710} & 0.9710        & 0.9660        & 0.9685          \\
Slovenian             & 0.9470        & 0.9380        & \textbf{0.9425} & 0.9450        & 0.9320        & 0.9385          \\
Spanish               & 0.9820        & 0.9810        & \textbf{0.9815} & 0.9810        & 0.9780        & 0.9795          \\
Swedish               & 0.9730        & 0.9700        & 0.9715          & 0.9730        & 0.9740        & \textbf{0.9735} \\
Thai                  & 0.9240        & 0.9150        & \textbf{0.9195} & 0.9130        & 0.8980        & 0.9055          \\
Turkish               & 0.8780        & 0.8820        & 0.8800          & 0.8910        & 0.8830        & \textbf{0.8870} \\
Ukrainian             & 0.9270        & 0.9190        & \underline{0.9230}    & 0.9260        & 0.9200        & \underline{0.9230}    \\
Urdu                  & 0.8830        & 0.8870        & 0.8850          & 0.8880        & 0.8840        & \textbf{0.8860} \\
Vietnamese            & 0.9700        & 0.9690        & \textbf{0.9695} & 0.9680        & 0.9640        & 0.9660       \\
\bottomrule
\end{tabular}
\end{table}

As discussed earlier, we implemented the student-teacher architecture and used a collection of parallel datasets, totalling over 128 million sentence pairs, as training data to train multi-lingual sentence encoders based on XLM-R and mBERT. The final checkpoints of the sentence encoders (CT-XLMR-SE and CT-mBERT-SE) were evaluated to assess: (i) how well they have learned the teacher's vector space and (ii) their multi-lingual capacity. The results from sentence encoding tasks are provided in Table \ref{eval-2} and Table \ref{vanilla}, and those from sentence matching tasks are summarized in Table \ref{eval-1}.

Table \ref{eval-2} presents results from sentence encoding tasks on real-world crisis-related human-labelled social media texts. The evaluated models include SBERT (Sentence Transformer's top-performing model: all-mpnet-base-v2), CT (CrisisTransformers), CT-XLMR-SE, and CT-mBERT-SE. All of these models utilized mean pooling over the token embeddings of an input sequence, considering the attention mask. Across all 18 datasets, our multi-lingual models consistently outperformed SBERT, whose $D_{avg}=$ 0.6374. CT-XLMR-SE with $D_{avg}=$ 0.7255 exhibited a 13.79\% improvement over SBERT, and CT-mBERT-SE with $D_{avg}=$ 0.7280 showed a 14.17\% improvement. We also present the performance of vanilla XLM-R and mBERT on sentence encoding tasks in Table \ref{vanilla}. The results show that sentence embeddings generated by these models lack semantic richness, aligning with findings in the existing literature that transformer-based pre-trained models, out-of-the-box, do not produce semantically meaningful sentence embeddings (\cite{reimers2019sentence, lamsal2023crisistransformers}).

Table \ref{eval-1} provides a detailed overview of the performance of the student models across 52 languages for sentence matching tasks on the TED2020 test sets. The evaluation metrics include accuracy scores for English to translated language task (en-t) and translated language to English task (t-en) and the average for each language. Out of 52 languages, CT-XLMR-SE performed best in 32 languages on average accuracy, while CT-mBERT-SE excelled in 18 languages. In 2 languages, both models performed equally. The results show variation among models in their performance across languages. Portuguese (Brazil), Spanish, Macedonian, Dutch, Portuguese, German, Serbian, and Slovak exhibit better results with CT-XLMR-SE. However, for languages such as Swedish, Romanian, and Croatian, CT-mBERT-SE surpasses CT-XLMR-SE. For Galician and Ukrainian, both models performed equally. CT-XLMR-SE achieves accuracy above 0.95 across 27 languages, while CT-mBERT-SE achieves this across 28 languages. Seven languages fall within the [0.8, 0.9) accuracy bracket for CT-XLMR-SE, and five languages for CT-mBERT-SE. The only outlier is CT-mBERT-SE, which achieved a score of 0.3055 for Kurdish (Sorani), placing it as the only language below the [0.8, 0.9) bracket. In contrast, CT-XLMR-SE achieved a score of 0.8890 for the same language. Overall, European languages tend to attain higher scores compared to languages from other regions. This trend could be attributed to the availability of more training data for European languages in the parallel datasets considered by this study.

\begin{table}[ht]
    \centering
    \caption{An originally posted crisis-related tweet and its translated versions.}
    \label{translated-pairs}
    \begin{tabular}{p{1.7cm}|p{13.3cm}}
    \toprule
    \textbf{Language} &  \textbf{Sentence}\\
    \midrule
    \midrule
\multicolumn{2}{c}{\textit{originally posted on Twitter$^{src}$}} \\
English   &  Our state is grappling with severe weather impacts right now, including Melbourne’s south east on the back of that terrible storm. I will keep you updated on federal recovery support and please stay safe everyone. \\
\midrule
\multicolumn{2}{c}{\textit{translated}} \\
Chinese (Simplified) & \begin{CJK*}{UTF8}{gbsn}我们的州目前正在努力应对恶劣天气的影响，包括遭受这场可怕风暴影响的墨尔本东南部。 我将随时向您通报联邦恢复支持的最新情况，请大家注意安全。\end{CJK*} \\
German & Unser Staat kämpft derzeit mit den Auswirkungen schwerer Unwetter, auch im Südosten Melbournes aufgrund dieses schrecklichen Sturms. Ich werde Sie über die Wiederaufbauunterstützung des Bundes auf dem Laufenden halten und bitte bleiben Sie alle gesund. \\
Italian & Il nostro Stato è alle prese con gravi conseguenze meteorologiche in questo momento, compreso il sud-est di Melbourne, a causa di quella terribile tempesta. Vi terrò aggiornati sul supporto federale per la ripresa e per favore state tutti al sicuro.\\
Japanese & \begin{CJK}{UTF8}{min}私たちの州は現在、あのひどい嵐の影響でメルボルンの南東部を含む、厳しい気象の影響と闘っています。 連邦政府の復興支援に関する最新情報をお知らせしますので、皆様ご安全にお過ごしください。\end{CJK} \\
Spanish  & Nuestro estado está lidiando con impactos climáticos severos en este momento, incluido el sureste de Melbourne a raíz de esa terrible tormenta. Los mantendré informados sobre el apoyo federal a la recuperación y por favor manténganse a salvo todos. \\
\bottomrule
    \end{tabular}
$^{src}$\url{https://twitter.com/ClareONeilMP/status/1757517855694954944}

\vspace{15pt}

\caption{Cosine similarities amongst the sentences listed in Table \ref{translated-pairs} (encoded by CT-XLMR-SE).}
\label{translated-pairs-cosine}
\begin{tabular}{l|cccccc}
         & \textbf{Chinese} & \textbf{English} & \textbf{German} & \textbf{Italian} & \textbf{Japanese} & \textbf{Spanish} \\
\midrule
\textbf{Chinese}  & --       & 0.934   & 0.931  & 0.930   & 0.939    & 0.937   \\
\textbf{English}  & 0.934   & --       & 0.943  & 0.969   & 0.944    & 0.970   \\
\textbf{German}   & 0.931   & 0.943   & --      & 0.956   & 0.936    & 0.947   \\
\textbf{Italian}  & 0.930   & 0.969   & 0.956  & --       & 0.959    & 0.970   \\
\textbf{Japanese} & 0.939   & 0.944   & 0.936  & 0.959   & --        & 0.952   \\
\textbf{Spanish}  & 0.937   & 0.970   & 0.947  & 0.970   & 0.952    & --      
\end{tabular}
    \end{table}

The mean squared error (MSE) loss used in the training is based on the deviation of student's embeddings from teacher's embeddings. Since the student is approximating the embedding space of the teacher, it does not surpass the teacher in tasks involving English texts. The teacher's specialized training on English texts gives it a more nuanced understanding of the language, allowing it to perform well in English tasks. However, the strength of the student model lies in its ability to process multiple languages. Sentence Transformers also offer multi-lingual versions of several of their pre-trained embedding models utilizing the same student-teacher embeddings deviation-based loss. Both our student models surpass Sentence Transformers' top-performing model, \textit{all-mpnet-base-v2}, in English sentence encoding tasks while maintaining multi-lingual capacity. As the second part of the training objective, i.e. $(\mathcal{CT}(en_j) - \mathcal{S}(t_j))$, involves aligning the student's non-English embeddings with the teacher's English embeddings, the student produces similar embeddings for translated pairs of sentences, which is evident with results and examples provided in Tables \ref{eval-1}, \ref{translated-pairs} and \ref{translated-pairs-cosine}. Table \ref{translated-pairs} lists an originally posted tweet and its translated (with Google Translate) versions, and Table \ref{translated-pairs-cosine} provides cosine similarities amongst them. Given that our student models replicate the embedding space of CrisisTransformers and surpass Sentence Transformers by $>13\%$, it is reasonable to expect that they perform well compared to Sentence Transformers' multi-lingual models. However, further study is required to quantify how well these models perform on real-world non-English crisis-related social media texts.

Multilingual embedding models are critical within crisis informatics. Areas where the proposed student models can be applied include semantic search, clustering, and topic modeling. Semantic search powered by multilingual embedding models enables the identification of semantically related content, such as matching help requests with relevant offers (\cite{dutt2019utilizing}), thus facilitating efficient allocation of resources during crises. For instance, if a first responding agency needs to filter through a large amount of social media messages, semantic search can quickly help identify messages indicating urgent needs, such as evacuation requests from a building collapse. By comparing the embeddings of search phrases with those of incoming social media streams in multiple languages, relevant messages can be efficiently retrieved and prioritized. For example, a similarity search with embeddings of any sentences in Table \ref{translated-pairs} with a similarity threshold of 0.9 would return remaining sentences as they are literal translations of each other. Moreover, multilingual embeddings are instrumental in clustering social media messages, offering benefits throughout the disaster management cycle. Clustering helps in organizing similar messages together, aiding in the identification of emerging trends, hotspot areas, or critical needs. This assists decision-makers in allocating resources and planning response strategies effectively. Also, neural topic models (\cite{grootendorst2022bertopic}) can use the cross-lingual sentence embeddings to perform topic modeling without language restrictions. These models can uncover underlying themes and topics within crisis-related social media data, providing insights into evolving situations and public concerns across diverse linguistic contexts.

\section{Conclusion}
In this study, we introduced two multi-lingual sentence encoders (CT-XLMR-SE and CT-mBERT-SE) designed for embedding crisis-related social media texts. Both models were trained as students in student-teacher networks, with CrisisTransformers' (mono-lingual) sentence encoder serving as the common teacher. The training process utilized a large-scale dataset comprising 10 different parallel datasets, totalling over 128 million sentence pairs (e.g., en–es, en–de, en–fr, etc.) across 52 languages. The proposed models underwent evaluation through sentence encoding tasks to assess how well they approximated the teacher's vector space and sentence matching tasks to evaluate their multi-lingual capabilities. Results from both tasks demonstrate that the models generalize well to the languages introduced during training and mimic CrisisTransformers' sentence encoder's embedding space. XLM-R and mBERT, upon which the proposed multi-lingual sentence encoders are based, were originally trained on extensive text corpora containing 100 and 104 languages, respectively. We performed additional training to align these models with CrisisTransformers' sentence encoder's vector space, concentrating on 52 specific languages.

\textbf{Advancing the area:} Our objective is to expand language coverage, including low-resource languages. There is increasing interest in distillation techniques to create small yet effective models from large pre-trained ones. This presents an exciting opportunity to distill knowledge from large models for generating state-of-the-art dense embeddings suitable for real-time processing of crisis-related social media texts. Additionally, we aim to integrate CrisisTransformers with distilled knowledge from large language models to enhance the quality of embeddings. Moreover, the crisis informatics domain lacks translation pairs to train a similar student-teacher architecture, suggesting a potential area for future exploration. Could crisis-specific translation pairs potentially outperform crisis-specific or general-purpose pre-trained models that are subsequently fine-tuned on parallel data from the general domain? This requires further investigation.

\section{Acknowledgements}
This study is supported by the Melbourne Research Scholarship from the University of Melbourne, Australia. This research was undertaken using the LIEF HPC-GPGPU Facility hosted at the University of Melbourne, which was established with the assistance of LIEF Grant LE170100200.

\printbibliography[heading=bibliography]

\end{document}